\documentclass[letterpaper, 10 pt, conference]{ieeeconf}  

\IEEEoverridecommandlockouts                              

\usepackage{graphicx} 
\usepackage{amsmath} 
\usepackage{amssymb}  
\usepackage{amsfonts}
\usepackage{xcolor}
\usepackage{epsfig}
\usepackage{graphicx}
\usepackage{subfigure}
\usepackage{booktabs} 
\usepackage{epstopdf} 
\usepackage[linesnumbered,ruled,vlined]{algorithm2e}
\SetNlSty{bfseries}{\color{black}}{}
\usepackage{algorithmic} 
\usepackage[colorinlistoftodos]{todonotes}
\title{\LARGE \bf
Learning Robot Exploration Strategy with 4D Point-Clouds-like \\ Information as Observations
}

\author{Zhaoting Li$^{1}$, Tingguang Li$^2$, Jiankun Wang$^{1*}$ and Max Q.-H. Meng$^{3*}$, \emph{Fellow, IEEE}
\thanks{*This work is supported by Shenzhen Key Laboratory of Robotics Perception and Intelligence (ZDSYS20200810171800001), Southern University of Science and Technology, Shenzhen 518055, China. (\emph{Corresponding author: Jiankun Wang, Max Q.-H. Meng}).}
\thanks{$^{1}$Zhaoting Li and Jiankun Wang are with the Department of Electronic and Electrical Engineering of the Southern University of Science and Technology in Shenzhen, China,
        {\tt\small \{lizt3@mail.,wangjk@\}sustech.edu.cn}}%
\thanks{$^{2}$Tingguang Li is with Tencent Robotics X Laboratory in Shenzhen, China. {\tt\small teaganli@tencent.com}}%
\thanks{$^{3}$Max Q.-H. Meng is with the Department of Electronic and Electrical Engineering of the Southern University of Science and Technology in Shenzhen, China, on leave from the Department of Electronic Engineering, The Chinese University of Hong Kong, Hong Kong, and also with the Shenzhen Research Institute of the Chinese University of Hong Kong in Shenzhen, China}%
}

\begin{document}

\maketitle
\thispagestyle{empty} 

\begin{abstract}

Being able to explore unknown environments is a requirement for fully autonomous robots.
Many learning-based methods have been proposed to learn an exploration strategy. 
In the frontier-based exploration, learning algorithms tend to learn the optimal or near-optimal frontier to explore. 
Most of these methods represent the environments as fixed size images and take these as inputs to neural networks. 
However, the size of environments is usually unknown, which makes these methods fail to generalize to real world scenarios.
To address this issue, we present a novel state representation method based on 4D point-clouds-like information, including the locations, frontier, and distance information.
We also design a neural network that can process these 4D point-clouds-like information and generate the estimated value for each frontier.
Then this neural network is trained using the typical reinforcement learning framework.
We test the performance of our proposed method by comparing it with other five methods and test its scalability on a map that is much larger than maps in the training set.  
The experiment results demonstrate that our proposed method needs shorter average traveling distances to explore whole environments and can be adopted in maps with arbitrarily sizes.
\end{abstract}

\section{Introduction}

Robot exploration problem is defined as making a robot or multi-robots explore unknown cluttered environments (i.e., office environment, forests, ruins, etc.) autonomously with specific goals. 
The goal can be classified as: (1) Maximizing the knowledge of the unknown environments, i.e., acquiring a map of the world \cite{thrun2002probabilistic}, \cite{28_background_info}. (2) Searching a static or moving object without prior information about the world. 
While solving the exploration problems with the second goal can combine the prior information of the target object, such as the semantic information, it also needs to fulfill the first goal. 

The frontier-based methods \cite{21_yamauchi1997frontier}, \cite{22_gonzalez2002navigation} have been widely used to solve the robot exploration problem. 
\cite{21_yamauchi1997frontier} adopts a greedy strategy, which may lead to an inefficient overall path.
Many approaches have been proposed to consider more performance metrics (i.e. information gain, etc.) \cite{22_gonzalez2002navigation}, \cite{23_bourgault2002information}, \cite{33_basilico2011exploration}. 
However, these approaches are designed and evaluated in a limited number of environments. Therefore, they may fail to generalize to other environments whose layouts are different.

\begin{figure}[t]
	\centering
	\includegraphics[scale=0.2]{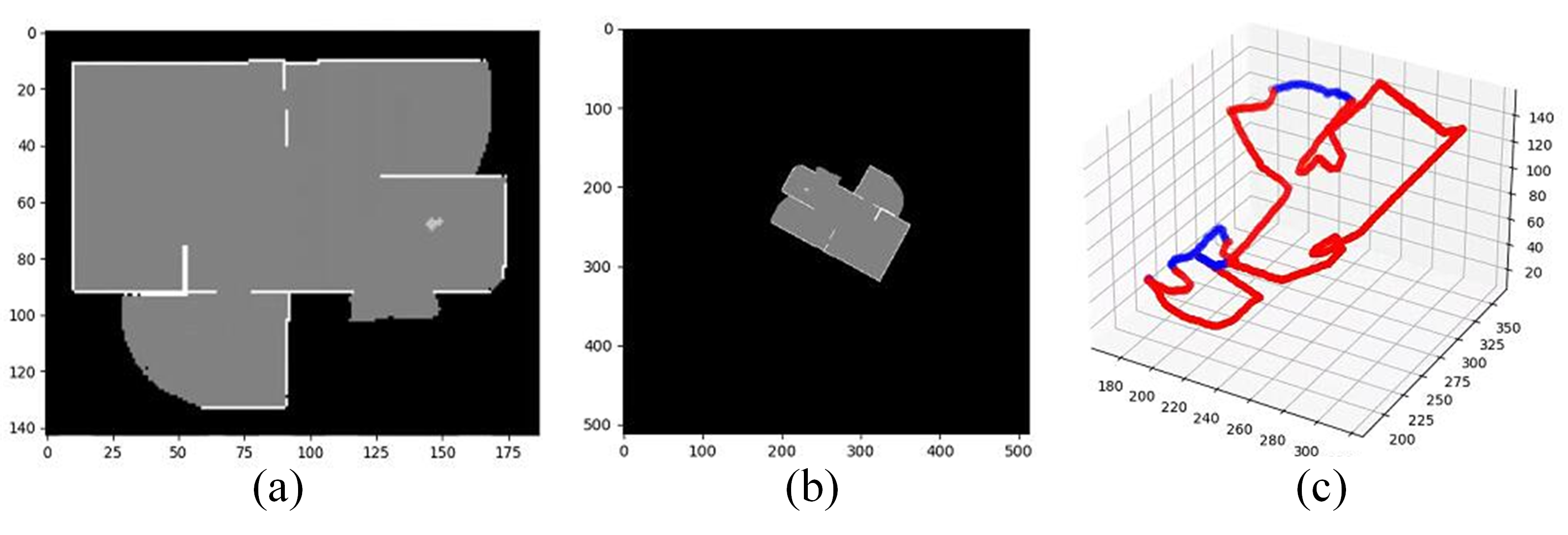}
	\caption{The example of 4D point-clouds-like information. (a) shows a map example in $HouseExpo$, where the black, gray, and white areas denote unknown space, free space, and obstacle space, respectively. (b) shows the global map in our algorithm, where the map is obtained by homogeneous transformations. The map's center is the robot start location, and the map's x-coordinate is the same as the robot start orientation. (c) shows the 4D point-clouds-like information generated based on the global map. The x, y, z coordinate denote x location, y location and distance information, respectively. The color denotes frontier information, where red denotes obstacles and blue denotes frontiers.}
	\label{fig:label_pointcloud}
\end{figure}

Compared with these classical methods, machine learning-based methods exhibit the advantages of learning from various data. 
Deep Reinforcement Learning (DRL), which uses a neural network to approximate the agent's policy during the interactions with the environments \cite{sutton2018reinforcement}, gains more and more attention in the application of games \cite{DQN2015}, robotics \cite{rl_env_robo}, etc.
When applied to robot exploration problems, most research works \cite{MaxM2018}, \cite{31_chen2019self} {design the state space as the form of image} and use Convolution Neural Networks (CNN).
For example, in \cite{MaxM2018}, CNN is utilized as a mapping from the local observation of the robot to the next optimal action.

However, the size of CNN's input images is fixed, which results in the following limitations: 
(1) If input images represent the global information, the size of the overall explored map needs to be pre-defined to prevent input images from failing to fully containing all the map information. 
(2) If input images represent the local information, which fails to convey all the state information, the recurrent neural networks (RNN) or memory networks need to be adopted. 
Unfortunately, the robot exploration problem in this formulation requires relatively long-term planning, which is still a tough problem that has not been perfectly solved.

In this paper, to deal with the aforementioned problems, we present a novel state representation method, which relies on 4D point-clouds-like information of variable size. 
These information have the same data structure as point clouds and consists of 2D points' location information, and the corresponding 1D frontier and 1D distance information, as shown in Fig.\ref{fig:label_pointcloud}.
We also designs the corresponding training framework, which bases on the deep Q-Learning method with variable action space.
By replacing the image observation with 4D point-clouds-like information, our proposed exploration model can deal with unknown maps of arbitrary size.
Based on dynamic graph CNN (DGCNN) \cite{wang2019dynamic}, which is one of the typical neural network structure that process point clouds information, our proposed neural network takes 4D point-clouds-like information as input and outputs the expected value of each frontier, which can guide the robot to the frontier with the highest value.
This neural network is trained in a way similar to DQN in the $HouseExpo$ environment \cite{li2019houseexpo}, which is a fast exploration simulation platform that includes data of many 2D indoor layouts. The experiment shows that our exploration model can achieve a relatively good performance, compared with the baseline in \cite{li2019houseexpo}, state-of-the-art in \cite{Weight2019}, classical methods in \cite{21_yamauchi1997frontier}, \cite{22_gonzalez2002navigation} and a random method.

\subsection{Original Contributions}
The contributions of this paper are threefold. 
First, we propose a novel state representation method using 4D point-clouds-like information to solve the aforementioned problems in Section \ref{sec:relatedwork}.
Although point clouds have been utilized in motion planning and navigation (\cite{27_2Dpointclouds}, \cite{30_ObstacleResp}), our work is different from these two papers in two main parts: 
(1) We use point clouds to represent the global information while they represent the local observation.
(2) Our action space is to select a frontier point from the frontier set of variable size, while their action space contains control commands.
Second, we design the corresponding state function based on DGCNN \cite{wang2019dynamic}, and the training framework based on DQN \cite{DQN2015}. 
The novelty is that our action space's size is variable, which makes our neural network converge in a faster way.
Third, we demonstrate the performance of the proposed method on a wide variety of environments, which the model has not seen before, and includes maps whose size is much larger than maps in the training set. 

The remainder of this paper is organized as follows. 
We first introduce the related work in Section \ref{sec:relatedwork}.
Then we formulate the frontier-based robot exploration problem and DRL exploration problem in Section \ref{sec:formulation}. 
After that, the framework of our proposed method are detailed in Section \ref{sec:alg}.
In Section \ref{sec:exp}, we demonstrate the performance of our proposed method through a series of simulation experiments. 
At last, we conclude the work of this paper and discuss directions for future work in section \ref{sec:conclude}.

\section{Related Work}
\label{sec:relatedwork}
In \cite{21_yamauchi1997frontier}, the classical frontier method is defined, where an occupancy map is utilized in which each cell is placed into one of three classes: open, unknown and occupied.
Then frontiers are defined as the boundaries between open areas and unknown areas. 
The robot can constantly gain new information about the world by moving to successive frontiers, while the problem of selecting which frontiers at a specific stage remains to be solved. 
Therefore, in a frontier-based setting, solving the exploration problem is equivalent to finding an efficient exploration strategy that can determine the optimal frontier for the robot to explore.  
A greedy exploration strategy is utilized in \cite{21_yamauchi1997frontier} to select the nearest unvisited, accessible frontiers. 
The experiment results in that paper show that the greedy strategy is short-sighted and can waste lots of time, especially when missing a nearby frontier that will disappear at once if selected (this case is illustrated in the experiment part).

Many DRL techniques have been applied into the robot exploration problem in several previous works.
In a typical DRL framework \cite{sutton2018reinforcement}, the agent interacts with the environment by taking actions and receiving rewards from the environment. 
Through this trial-and-error manner, the agent can learn an optimal policy eventually.
In \cite{MLiu2016}, a CNN network is trained under the DQN framework with RGB-D sensor images as input to make the robot learn obstacle avoidance ability during exploration. 
Although avoiding obstacles is important, this paper does not apply DRL to learn the exploration strategy.
The work in \cite{Weight2019} combines frontier-based exploration with DRL to learn an exploration strategy directly. 
The state information includes the global occupancy map, robot locations and frontiers, while the action is to output the weight of a cost function that evaluates the goodness of each frontier. 
The cost function includes distance and information gain. 
By adjusting the weight, the relative importance of each term can be changed. 
However, the terms of the cost function rely on human knowledge and may not be applicable in other situations. 
In \cite{2_li2019deep}, the state space is similar to the one in \cite{Weight2019}, while the action is to select a point from the global map.
However, the map size can vary dramatically from one environment to the next. 
It losses generality when setting the maximum map size before exploring the environments. 

In \cite{MaxM2018} and \cite{31_chen2019self}, a local map, which is centered at the robot's current location, is extracted from the global map to represent current state information.
By using a local map, \cite{MaxM2018} trains the robot to select actions in ``turn left, turn right, move forward'', while \cite{31_chen2019self} learns to select points in the free space of the local map. 
Local map being state space can eliminate the limitation of global map size, but the current local map fails to contain all the information. 
In \cite{MaxM2018}, the robot tends to get trapped in an explored room when there is no frontier in the local map, because the robot has no idea where the frontiers are.
The training process in \cite{31_chen2019self} needs to drive the robot to the nearest frontier when the local map contains no frontier information, although a RNN network is integrated into their framework. 
This human intervention adopts a greedy strategy and can not guarantee an optimal or near-optimal solution. 
When utilizing local observations, the robot exploration problem requires the DRL approach to have a long-term memory.
Neural map in \cite{17_parisotto2018neural} is proposed to tackle simple memory architecture problems in DRL.
Besides, Neural SLAM in \cite{13_zhang2017neural} embeds traditional SLAM into attention-based external memory architecture.
However, the memory architectures in \cite{17_parisotto2018neural} and \cite{13_zhang2017neural} are based on the fixed size of the global map.
It is difficult to be applied to unknown environments whose size may be quite large compared with maps in training sets.
Unlike the aforementioned methods, our method uses the 4D point-clouds-like information to represent the global state information, which does not suffer from both the map size limitation and the simple memory problem. 
As far as we know, our method is the first to apply point clouds to robot exploration problems. 
Therefore, we also design a respective DRL training framework to map the exploration strategy directly from point clouds.

\section{Problem Formulation}
\label{sec:formulation}

Our work aims to develop and train a neural network that can take 4D point-clouds-like information as input and generate efficient policy to guide the exploration process of a robot equipped with a laser scanner. 
The network should take into account the information about explored areas, occupied areas, and unexplored frontiers. 
In this paper, the robot exploration problem is to make a robot equipped with a limited-range laser scanner explore unknown environments autonomously.

\subsection{Frontier-based Robot Exploration Problem}
\label{sec:formulation_A}
In the robot exploration problem, a 2D occupancy map is most frequently adopted to store the explored environment information. 
Define the explored 2D occupancy map at step ${t}$ as ${M_t}$.
Each grid in ${M_t}$ can be classified into the following three states: free grid ${M_t}$, occupied grid ${E_t}$, and unknown grid ${U_t}$. 
According to \cite{21_yamauchi1997frontier}, frontiers ${F_t}$ are defined as the boundaries between the free space ${E_t}$ and unknown space ${U_t}$.
Many existing DRL exploration frameworks learn a mapping from ${M_t}$ and ${F_t}$ to robot movement commands which can avoid obstacles and navigate to specific locations.
Although this end-to-end framework has a simple structure, it is difficult to train.
Instead, our method learns a policy network that can directly determine which frontier to explore, which is similar to \cite{Weight2019} and \cite{2_li2019deep}.
At step $t$, a target frontier is selected from ${F_t}$ based on an exploration strategy and current explored map ${M_t}$. 
Once the change of the explored map is larger than the threshold, the robot is stopped, and the explored map at step $t+1$ ${M_{t+1}}$ is obtained.
By moving to selected frontiers constantly, the robot will explore more unknown areas until no accessible frontier exists. 

Because ${M_t}$ can be represented as an image, it is commonly used to directly represent the state information.
As explained in Section \ref{sec:relatedwork}, a novel state representation method with 4D point-clouds-like information is proposed instead.
The 4D point-clouds-like information at step $t$ is defined as a 4-dimensional point cloud set with $n$ points, denoted by $X_t = \{ x_1^t, ..., x_n^t \} \subset \mathbb{R}^4$.
Each point contains 4D coordinates $x_i^t = \left( x_i, y_i, b_i, d_i\right)$, where $x_i, y_i$ denotes the location of the point, $d_i$ denotes the distance from the point to the robot location without collision, $b_i \in \{ 0, 1\}$ denotes whether point $(x_i, y_i)$ in $M_t$ belongs to frontier or not.

\begin{figure}[t]
	\centering
	\includegraphics[scale=0.53]{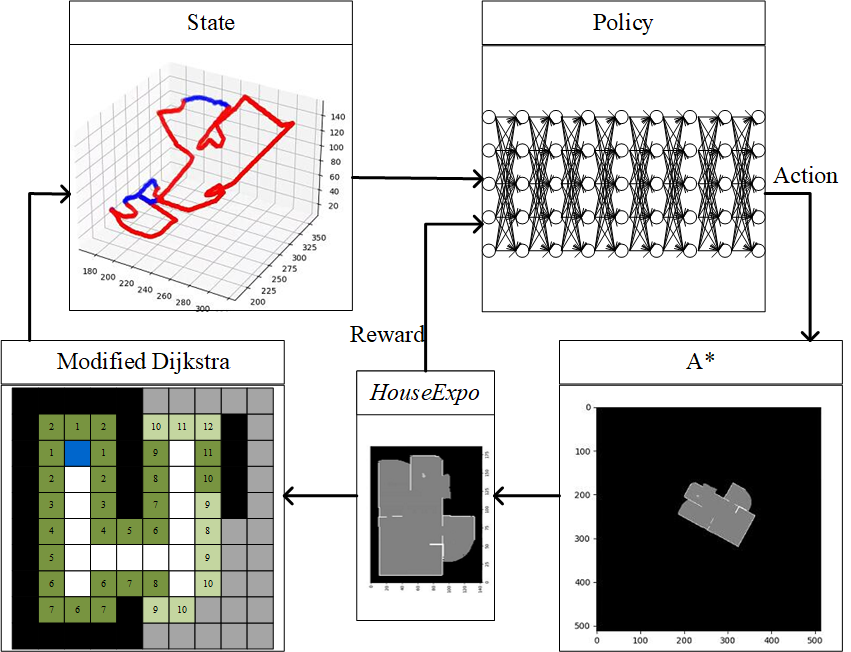}
	\caption{The framework of the proposed method, which consists of five components: (a) A simulator adopted from $HouseExpo$ \cite{li2019houseexpo}, which receives and executes a robot movement command and outputs the new map in $HouseExpo$ coordinate; (b) The modified Dijkstra algorithm to extract the contour of the free space; (c) The state represented by 4D point-clouds-like information; (d) The policy network which processes the state information and estimates the value of frontiers; (e) The A* algorithm that finds a path connecting the current robot location to the goal frontier and a simple path tracking algorithm that generates the corresponding robot movement commands.}
	\label{fig:label_framework}
\end{figure}

\subsection{DRL exploration formulation}
\label{sec:DRL_formulation}
The robot exploration problem can be formulated as a Markov Decision Process (MDP), which can be modeled as a tuple $(\mathcal{S}, \mathcal{A}, \mathcal{T}, \mathcal{R}, \gamma)$.
The state space ${\mathcal{S}}$ at step $t$ is defined by 4D point cloud $X_t$, which can be divided into frontier set $F_t$ and obstacle set $O_t$.
The action space ${\mathcal{A}_t}$ at step $t$ is the frontier set $F_t$, and the action is to select a point $f_t$ from $F_t$, which is the goal of the navigation module implemented by $A^{*}$ \cite{Hart1968Astar}.
When the robot take an action $f_t$ from the action space, the state $X_t$ will transit to state $X_{t+1}$ according to the stochastic state transition probability ${\mathcal{T}(X_t, f_t, X_{t+1})} = p(X_{t+1} | X_t, f_t)$.
Then the robot will receive an immediate reward $r_t = \mathcal{R}(X_t, f_t)$.
The discount factor $\gamma \in [0, 1]$ adjusts the relative importance of immediate and future reward.
The objective of DRL algorithms is to learn a policy $\pi (f_t | X_t)$ that can select actions to maximize the expected reward, which is defined as the accumulated $\gamma-$discounted rewards over time.

Because the action space varies according to the size of frontier set $F_t$, it is difficult to design a neural network that maps the state to the action directly.
The value-based RL is more suitable to this formulation. 
In value-based RL, a vector of action values, which are the expected rewards after taking actions in state $X_t$ under policy $\pi$, can be estimated by a deep Q network (DQN) $Q_{\pi}(X_t, f_t; \theta) = \mathbb{E}[\sum_{i=t}^{\infty}{\gamma}^{i-t}r_i | X_t, f_t]$, where $\theta$ are the parameters of the muti-layered neural network.
The optimal policy is to take action that has the highest action value:
$f_{t}^{*}=\mathop{\text{argmax}}_{f_t}Q_{\pi}(X_t,f_t; \theta)$.
DQN \cite{DQN2015} is a novel variant of Q-learning, which utilizes two key ideas: 
experience reply and target network. 
The DQN tends to overestimate action values, which can be tackled by double DQN in \cite{doubleDQN}.
Double DQN select an action the same as DQN selects, while estimate this action's value by the target network.

\section{Algorithm}
\label{sec:alg}

\begin{figure}[t]
	\centering
	\includegraphics[scale=0.3]{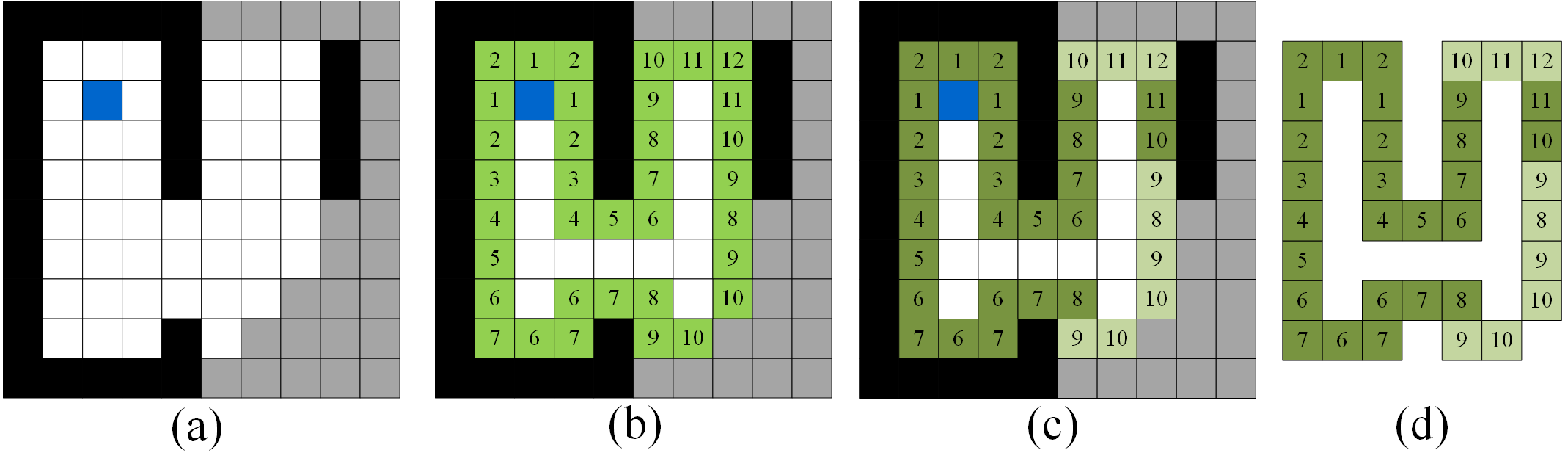}
	\caption{The illustration of the 4D point-clouds-like information generation process. (a) presents the map where the black, gray, white, blue points denote obstacles, unknown space, free space, and robot location. In (b), the contour of free space, which is denoted by green points, is generated by the modified Dijkstra algorithm. The number on each point indicates the distance from this point to the robot location. In (c), the points in the contour set are divided into frontier or obstacle sets, which are denoted in dark green and light green, respectively. In (d), the 4D point-clouds-like information are extracted from the image (c). The point clouds include location, frontier flag, and distance information.}
	\label{fig:label_illustration_h}
\end{figure}

In this section, we present the framework of our method and illustrate its key components in detail. 

\subsection{Framework Overview}
The typical DRL framework is adopted, where the robot interacts with the environment step by step to learn an exploration strategy. 
The environment in this framework is based on $HouseExpo$ \cite{li2019houseexpo}.
The state and action space of the original $HouseExpo$ environment is the local observation and robot movement commands, respectively. 
When incorporated into our framework, $HouseExpo$ receives a sequence of robot movement command and outputs the global map once the change of the explored map is larger than the threshold, which is detailed in Section \ref{sec:formulation_A}.
As shown in Fig. \ref{fig:label_framework}, the 4D point-clouds-like information can be obtained by a modified Dijkstra algorithm. 
After the policy network outputting the goal point, the $A^{*}$ algorithm is implemented to find the path connecting the robot location to the goal point. 
Then a simple path tracking algorithm is applied to generate the sequence of robot movement commands.

\subsection{Frontier Detection and Distance Computation}
\label{sec:frontier_dectect}
Computing the distances from the robot location to points in the frontier set without collision will be time-consuming if each distance is obtained by running the $A^{*}$ algorithm once. 
Instead, we modify the Dijkstra algorithm to detect frontiers and compute distance at the same time by sharing the search information.  
Denote the ``open list'' and ``close list'' as $L_o$ and $L_c$, respectively.
The open list contains points that need to be searched, while the close list contains points that have been searched.
Define the contour list as $L_f$, which contains the location and the cost of points that belong to frontier or obstacle.
Only points in the free space of map $M_t$ are walkable. 
The goal of this modified algorithm is to extract the contour of free space and obtain the distance information simultaneously.
As shown in Algorithm 1, the start point with the cost of zero, which is decided by the robot location, is added to $L_o$. 
While the open list is not empty, the algorithm repeats searching 8 points, denoted by $p_{near}$, adjacent to current point $p_{cur}$.
The differences from Dijkstra algorithms are: 
(1) If $p_{near}$ belongs to occupied or unknown space, add $p_{cur}$ to frontier list $L_f$, as shown in line 10 of Algorithm 1.
(2) Instead of stopping when the goal is found, the algorithm terminates until $L_o$ contains zero points.
After the algorithm ends, the contour list contains points that are frontiers or boundaries between free space and obstacle space.
Points in the contour list can be classified by their neighboring information into frontier or obstacle set, which is shown in Fig. \ref{fig:label_illustration_h}.

\begin{algorithm}[h]
\caption{Modified Dijkstra Algorithm}
\begin{algorithmic}[1]
\STATE $L_o \leftarrow \{ p_{start}\}, Cost( p_{start}) = 0;$ 
\WHILE {$L_o \neq \phi$}
\STATE $p_{cur} \leftarrow minCost(L_o)$;
\STATE $L_c \leftarrow L_c \cup p_{cur}, L_o \leftarrow L_o \backslash p_{cur}$;
\FOR{$p_{near}$ in 8 points adjacent to $p_{cur}$}
\STATE $cost_{near} = Cost(p_{cur}) + distance(p_{near}, p_{cur})$
\IF{$p_{near} \in L_c$}
\STATE continue;
\ELSIF{$p_{near}$ is not walkable}
\STATE $L_f \leftarrow L_f \cup p_{cur}$;
\ELSIF{$p_{near} \notin L_o$}
\STATE $L_o \leftarrow L_o \cup p_{near}$
\STATE $Cost(p_{near}) = cost_{near}$;
\ELSIF{$p_{near} \in L_o$ and $Cost(p_{near})>cost_{near}$}
\STATE $Cost(p_{near})=cost_{near}$
\ENDIF
\ENDFOR
\ENDWHILE 
\label{code:recentEnd}
\end{algorithmic}
\end{algorithm}

\begin{figure*}[t]
	
	\centering
	\includegraphics[scale=0.48]{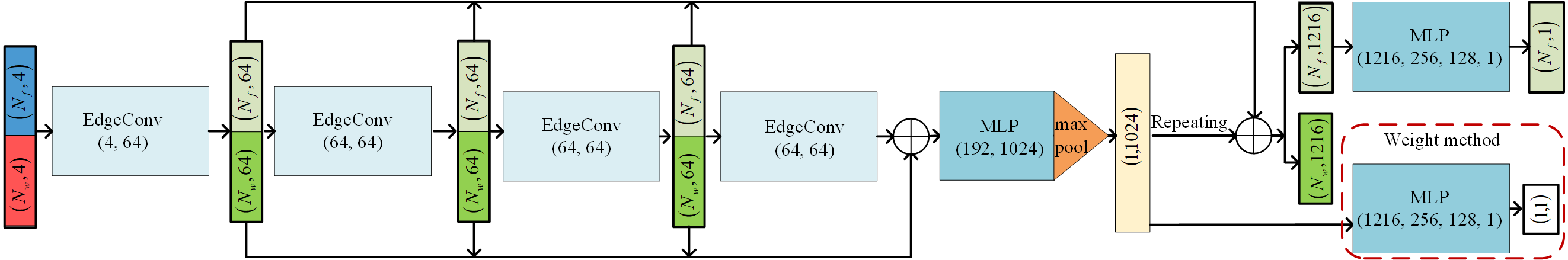}
	\caption{The architecture of our proposed neural network. The input point clouds are classified into two categories: frontier set denoted by dark blue and obstacle set denoted by red. The edge convolution operation $EdgeConv$ is denoted by a light blue block, which is used to extract local features around each point in the input set. The feature sets of frontiers and obstacles, which are generated by $EdgeConv$ operations, are denoted by light green and green. The MLP operation is to extract one point's feature by only considering this point's information. After several $EdgeConv$ operations and one MLP operation, a max-pooling operation is applied to generate the global information, which is shown in light yellow.}
	\label{fig:label_network_structure}
\end{figure*}

\subsection{Network with Point Clouds as Input}
\label{sec:DQN_architecture}
In this section, the architecture of the state-action value network with 4D point-clouds-like information as input is detailed.
The architecture is modified from DGCNN in Segmentation task \cite{wang2019dynamic}, which proposes edge convolution (EdgeConv) operation to extract the local information of point clouds.
The EdgeConv operation includes two steps for each point in the input set: (1) construct a local graph including the center point and its k-nearest neighbors; (2) apply convolution-like operations on edges which connect each neighbor to the center point.
The $(D_{in}, D_{out})$ EdgeConv operation takes the point set $(N, D_{in})$ as input and outputs the feature set $(N, D_{out})$, where $D_{in}$ and $D_{out}$ denote the dimension of input and output set, and $N$ denotes the number of points in the input set.
Different from DGCNN and other typical networks processing point cloud such as PointNet \cite{qi2017pointnet}, which have the same input and output point number, our network takes the frontier and obstacle set as input and only outputs the value of points from the frontier set.
The reason for this special treatment is to decrease the action space's size to make the network converge in a faster manner.

The network takes as input $N_f + N_w$ points at time step $t$, which includes $N_f$ frontier points and $N_w$ obstacle points, which are denoted as $F_t = \{ x_1^t, ..., x_{N_f}^t \}$ and $O_t = \{ x_{N_f+1}^t, ..., x_{N_f+N_w}^t \}$ respectively.
The output is a vector of estimated value of each action $Q_{\pi}(F_t, O_t, \cdot; \theta)$.
The network architecture contains multiple EdgeConv layers and multi-layer perceptron (mlp) layers.
At one EdgeConv layer, the feature set is extracted from the input set. 
Then all features in this edge feature set are aggregated to compute the output EdgeConv feature for each corresponding point.
At a mlp layer, the data of each point is operated on independently to obtain the information of one point, which is the same as the mlp in PointNet \cite{qi2017pointnet}.
After 4 EdgeConv layers, the outputs of all EdgeConv layers are aggregated and processed by a mlp layer to generate a 1D global descriptor, which encodes the global information of input points.
Then this 1D global descriptor is concatenated with the outputs of all EdgeConv layers.
After that, points that belong to the obstacle set are ignored, and the points from the frontier set are processed by three mlp layers to generate scores for each point.

\subsection{Learning framework based on DQN}
As described in Section \ref{sec:DRL_formulation}, the DQN is a neural network that for a given state $X_t$ outputs a vector of action values.
The network architecture of DQN is detailed in Section \ref{sec:DQN_architecture}, where the state $X_t$ in point clouds format contains frontier and obstacle sets.
Under a given policy $\pi(f_t|F_t,O_t)$, the true value of an action $f_t$ in state $X_t = \{F_t, O_t \}$ is: $Q_{\pi}(X_t, f_t; \theta) \equiv \mathbb{E}[\sum_{i=t}^{\infty}{\gamma}^{i-t}r_i | X_t, f_t]$. 
The target is to make the estimate from the frontier network converge to the true value. 
The parameters of the action-value function can be updated in a gradient decent way, after taking action $f_t$ in state $X_t$ and observing the next state $X_{t+1}$ and immediate reward $r_{t+1}$: 
\begin{eqnarray}\label{update_Q}
	\theta_{t+1} = \theta_t + \alpha (G_t - Q_{\pi}(X_t, f_t; \theta_t)),
\end{eqnarray}
where $\alpha$ is the learning rate and the target $G_t$ is defined as 
\begin{eqnarray}\label{target}
	G_t = r_{t+1}+\gamma Q_{\pi}(X_{t+1}, \mathop{\text{argmax}}_{f}Q(X_{t+1}, f; \theta_t); \theta_t^{'}),
\end{eqnarray}
where $\theta^{'}$ denotes the parameter of the target network, which is updated periodically by $\theta^{'} = \theta$.

To make the estimate from our network converge to the true value, a typical DQN framework \cite{DQN2015} can be adopted. 
For each step $t$, the tuple $(X_t, f_t, X_{t+1}, r_{t+1})$ is saved in a reply buffer.  
The parameters can be updated by equation \ref{update_Q} and \ref{target} given the tuple sampled from the reply buffer.

The reward signal in the DRL framework helps the robot know at a certain state $X_t$ whether an action $f_t$ is appropriate to take.
To make the robot able to explore unknown environments successfully, we define the following reward function:

\begin{equation}\label{reward_all}
r_t = r_{area}^t + r_{frontier}^t + r_{action}^t.
\end{equation}
The term $r_{area}^t$ equals the newly discovered area at time $t$, which is designed to encourage the robot to explore unknown areas.
The term $r_{action}^t$ provides a consistent penalization signal to the robot when a movement command is taken. 
This reward encourages the robot to explore unknown environments with a relatively short length of overall path.
The term $r_{frontier}^t$ is designed to guide the robot to reduce the number of frontiers, as defined in equation \ref{reward_f}:
\begin{equation}\label{reward_f}
r_{frontier}^t = \left\{
\begin{aligned}
 1,  N_{frontier}^t < N_{frontier}^{t-1}\\
 0,  N_{frontier}^t \geq N_{frontier}^{t-1}.
\end{aligned}
\right.
\end{equation}
where $N_{frontier}^t$ denotes the number of frontier group at time $t$.

\section{Training Details and Simulation Experiments}
\label{sec:exp}
In this section, we first detail the training process of our DRL agent. 
Then we validate the feasibility of our proposed method in robot exploration problem by two sets of experiments:
(1) a comparison with five other exploration methods, (2) a test of scalability where maps of larger size compared with training set are to be explored.

\subsection{Training Details}
\begin{figure}[t]
	\centering
	\includegraphics[scale=0.3]{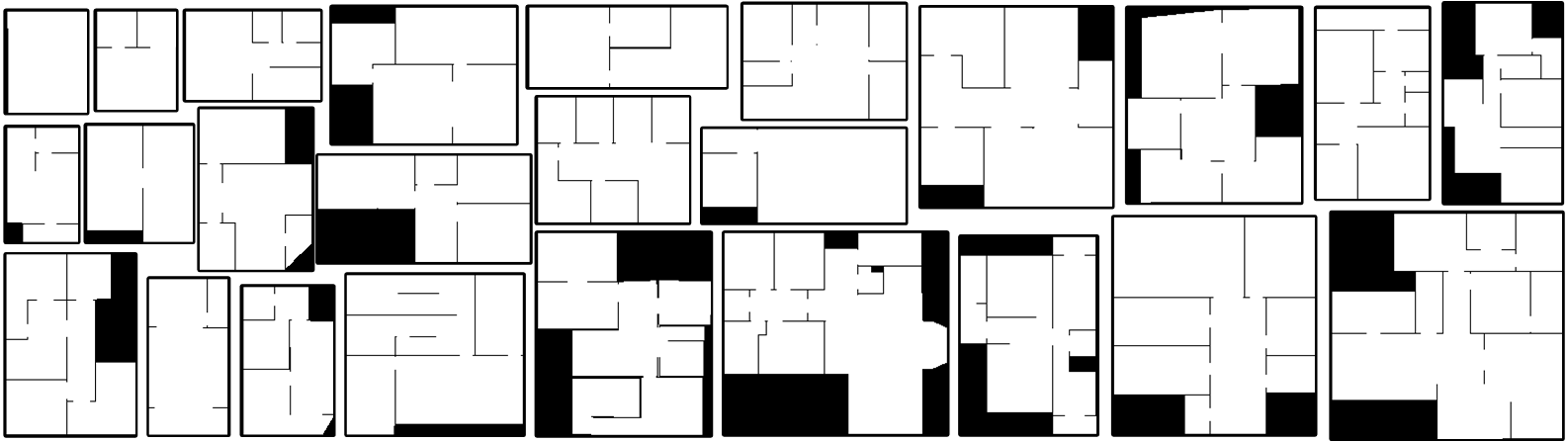}
	\caption{Some map samples from training set. The black and white pixels represent obstacle and free space respectively.}
	\label{fig:label_mapsample}
\end{figure}

To learn a general exploration strategy, the robot is trained in the $HouseExpo$ environment where 100 maps with different shapes and features are to be explored.
The robot is equipped with a 2m range laser scanner with a 180 degree field of view. 
The noise of laser range measurement is simulated by the Gaussian distribution with a mean of 0 and a standard deviation of 0.02m.
As an episode starts, a map is randomly selected from the training set, and the robot start pose, including the location and pose, is also set randomly.
In the beginning, the circle area centered at the start location is explored by making the robot rotate for 360 degree.
Then at each step, a goal frontier point is selected from the frontier set under the policy of our proposed method.
A* algorithm is applied to find a feasible path connecting the current robot location to the goal frontier. 
A simple path tracking algorithm is used to find the robot commands to follow the planned path: moving to the nearest unvisited path point $p_{near}$, and replanning the path if the distance between the robot and $p_{near}$ is larger than a fixed threshold.
An episode ends if the explored ratio of the whole map is over ${95\%}$.

\begin{table}[!htbp]
\caption{Parameters in Training}
\centering
\begin{tabular}{cccc}
    \toprule
    \multicolumn{2}{c}{HouseExpo} & \multicolumn{2}{c}{Training} \\
    \hline
    Laser range & $2m$ & Discount factor $\gamma$ & 0.99 \\
    \hline
    Laser field of view & $180^{\circ}$ & Target network update f & 4000 steps \\
    \hline
    Robot radius & $0.15m$ & Learning rate & 0.001 \\
    \hline
    Linear step length & $0.3m$ & Replay buffer size & 50000 \\
    \hline
    Angular step length & $15^{\circ}$ & Exploration rate $\epsilon$ & 15000\\
    \hline
    Meter2pixel & $16$ & Learning starts & 3000\\
    \bottomrule
\end{tabular}
\end{table}

Some training map samples are shown in Fig. \ref{fig:label_mapsample}. 
The largest size of a map in the training set is 256 by 256.
Because the size of state information $X_t$ changes at each time and batch update method requires point clouds with same size, currently, it is not realistic to train the model in a batch-update way.
In typical point clouds classification training process, the size of all the point clouds data are pre-processed to the same size.
However, these operations will change the original data's spatial information in our method.
Instead, for each step, the network parameters are updated 32 times with a batch size equal to 1.
Learning parameters are listed in Table 1. 
The training process is performed on a computer with Intel i7-7820X CPU and GeForce GTX 1080Ti GPU.
The training starts at 2000 steps and ends after 90000 update steps, which takes 72 hours.

\subsection{Comparison Study}
\begin{figure}[t]
	\centering
	\includegraphics[scale=0.65]{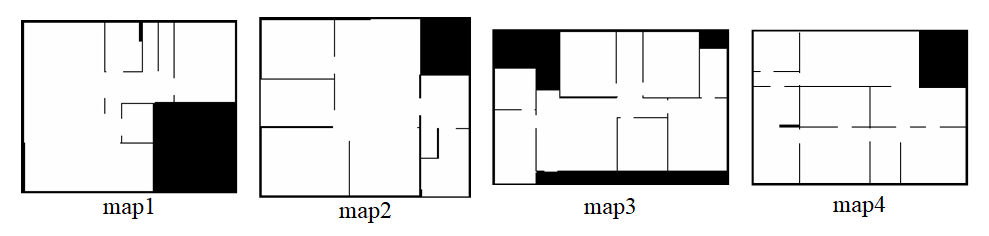}
	\caption{Four maps for testing. The size of map1, map2, map3 and map4 is (234, 191), (231, 200), (255, 174) and (235, 174), respectively. }
	\label{fig:test_data}
\end{figure}
\begin{figure}[t]
	\centering
	\includegraphics[scale=0.34]{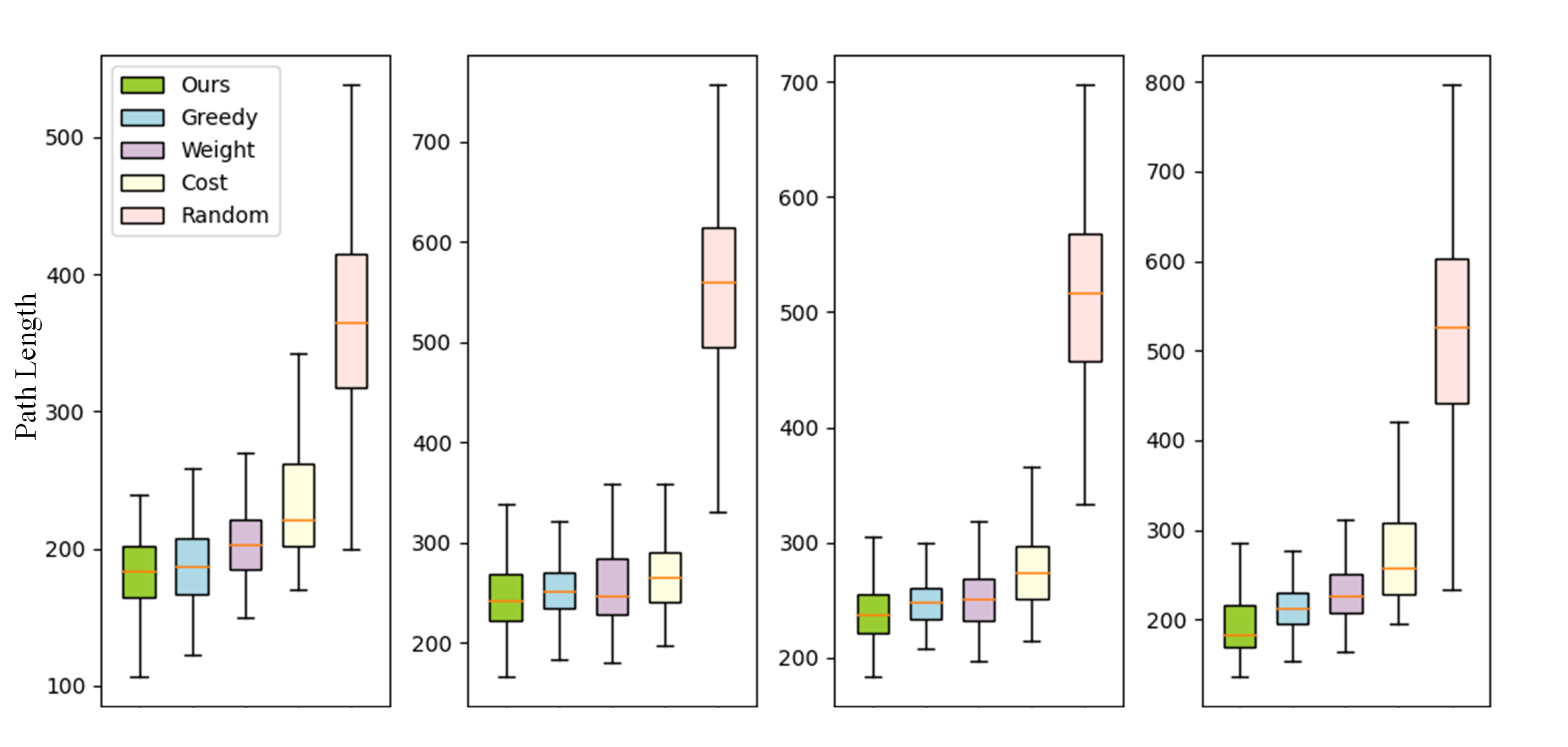}
	\caption{The path length's data of each method on four test maps. For each map, each method is tested for 100 times with the same randomly initialized robot start locations.}
	\label{fig:length_comparision}
\end{figure}

Besides our proposed method, we also test the performance of the weight tuning method in \cite{Weight2019}, a cost-based method in \cite{22_gonzalez2002navigation}, a method with greedy strategy in  \cite{21_yamauchi1997frontier}, a method utilizing a random policy and the baseline in \cite{li2019houseexpo}, which we denote as the weight method, cost method, greedy method, random method and baseline, respectively.
To compare the performance of different methods, we use 4 maps as a testing set, as shown in Fig. \ref{fig:test_data}.
For each test map, we conduct 100 trials by setting robot initial locations randomly for each trail.

The \emph{random method} selects a frontier point from the frontier set randomly.
The \emph{greedy method} chooses the nearest frontier point.
The \emph{baseline} utilizes a CNN which directly determines the robot movement commands by the current local observation.

The \emph{cost method} evaluates the scalar value of frontiers by a cost function considering distance and information gain information. 
\begin{equation}\label{cost_method}
cost = wd+(1-w)(1-g),
\end{equation}
where $w$ is the weight that adjusts the relative importance of costs. 
$d$ and $g$ denote the normalized distance and information gain of a frontier.
At each step, after obtaining the frontier set as detailed in Section \ref{sec:frontier_dectect}, the k-means method is adopted to cluster the points in the frontier set to find frontier centers. 
To reduce the runtime of computing information gains, we only compute the information gain for each frontier center.
The information gain is computed by the area of the unknown space to be explored if this frontier center is selected \cite{22_gonzalez2002navigation}.
The weight in the cost method is fixed to 0.5, which fails to be optimal in environments with different structures and features.

The \emph{weight method} can learn to adjust the value of the weight in equation \ref{cost_method} under the same training framework as our proposed method.
The structure of the neural network in weight method is presented in Fig. \ref{fig:label_network_structure}, which takes the 4D point-clouds-like information as input and outputs a scalar value.

\begin{figure}[t]
	\centering
	\includegraphics[scale=0.9]{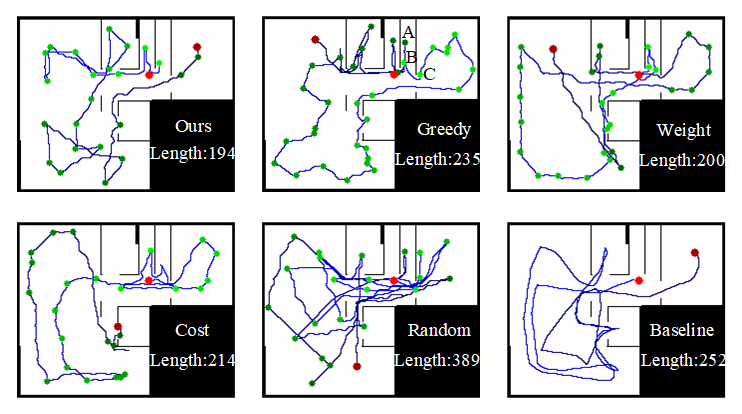}
	\caption{The representative exploration trials of six different methods on map1 with the same robot start locations and poses. The bright red denotes the start state and the dark red denotes the end state. The green point denotes the locations where the robot made decisions to choose which frontiers to explore. As the step increases, the brightness of green points becomes darker and darker. The baseline's result doesn't have green points because its action is to choose a robot movement command.}
	\label{fig:map_compare}
\end{figure}

We select the length of overall paths, which are recorded after $95\%$ of areas are explored, as the metric to evaluate the relative performance of the six exploration methods. 
The length of overall paths can indicate the efficiency of the exploration methods.
The box plot in Fig. \ref{fig:length_comparision} is utilized to visualize the path length's data of each exploration method on four test maps. 
The baseline is not considered here because the baseline sometimes fails to explore the whole environment, which will be explained later. 
Three values of this metric are used to analyze the experiment results: (1) the average, (2) the minimum value, (3) the variance.
Our proposed method has the minimum average length of overall paths for all four test maps.
The minimum length of the proposed method in each map is also smaller than other five methods.
This indicates that the exploration strategy of our proposed method is more effective and efficient than the other five methods.
The random method has the largest variance and average value for each map because it has more chances to sway between two or more frontier groups. 
That is why the overall path of the random method in Fig. \ref{fig:map_compare} is the most disorganized.
The weight method has a lower average and minimum value compared with the cost method in all test maps, due to the advantages of learning from rich data.
However, the weight method only adjusts a weight value to change the relative importance of distance and information gain. 
If the computation of the information gain is not accurate or another related cost exists, the weight method fails to fully demonstrate the advantages of learning.
Instead, our proposed method can learn useful information, including information gain, by learning to estimate the value of frontiers, which is the reason that our proposed method outperforms the weight method.

\begin{figure}[t]
	\centering
	\includegraphics[scale=0.38]{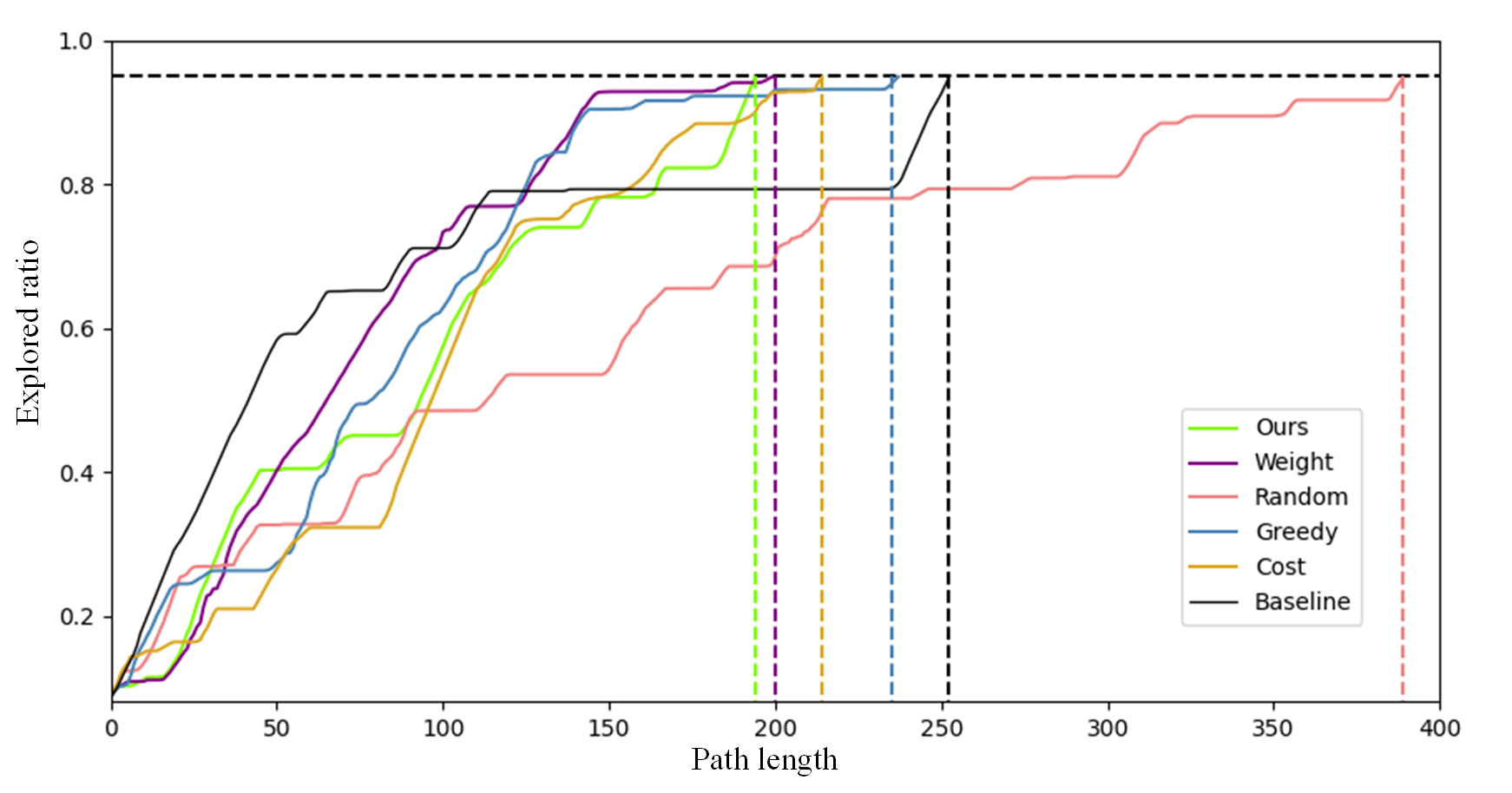}
	\caption{The ratio of the area explored and the area of the whole map with respect to the current path's length. The test map is map1 and the start locations is the same as Fig. \ref{fig:map_compare}.}
	\label{fig:label_explored_ratio}
\end{figure}

Fig. \ref{fig:map_compare} shows an example of the overall paths of six different methods when exploring the map1 with the same starting pose.
Each episode ended once the explored ratio was more than 0.95.
The proposed method explored the environment with the shortest path.
The explored ratio of the total map according to the length of the current path is shown in Fig. \ref{fig:label_explored_ratio}.
The greedy method's curve has a horizontal line when the explored ratio is near 0.95. 
This is because the greedy method missed a nearby frontier which would disappear at once if selected. 
However, the greedy method chose the nearest frontier instead, which made the robot travel to that missed frontier again and resulted in a longer path.
For example, in Fig. \ref{fig:map_compare}, the greedy method chose to explore the point C instead of the point A, when the robot was at point B. This decision made the robot travel to the point A later, thus making the overall path longer.
The baseline's curve also exhibits a horizontal line in Fig. \ref{fig:label_explored_ratio}, which is quite long.
The baseline's local observation contained no frontier information when the surroundings were all explored (in the left part of the environment shown in Fig. \ref{fig:map_compare}).
Therefore, at this situation, the baseline could only take ``random'' action (i.e. travelling along the wall) to find the existing frontiers, which would waste lots of travelling distances and may fail to explore the whole environment.

\subsection{Scalability Study}

\begin{figure}[t]
	\centering
	\includegraphics[scale=1.2]{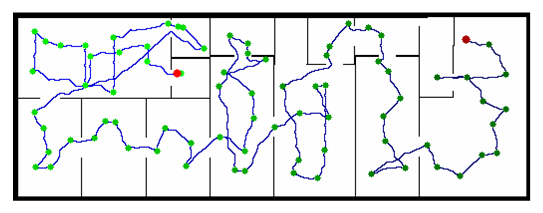}
	\caption{The result of the scalability test. The map size is (531, 201). The meaning of points' color is the same as Fig. \ref{fig:map_compare}}
	\label{fig:label_scalabilityTest}
\end{figure}

In this section, a map size of (531, 201) is used to test the performance of our proposed method in larger environments compared with maps in the training set. 
If the network's input is fixed size images, the map needs to be padded into a (531, 531) image, which is a low-efficient state representation way. 
Then a downscaling operation of the image is required to make the size of the image input the same as the requirement of the neural network, e.g. (256,256).
Although the neural network can process the state data by downscaling, the quality of the input data decreases.
Therefore, the network fails to work once the scaled input contains much less necessary information than the original image.
Fig. \ref{fig:label_scalabilityTest} presents the overall path generated by our method without downscaling the map size.
Our proposed method, which takes point clouds as input, has better robustness in maps with large scales because of the following two reasons.
First, we incorporate the distance information into point clouds, which can help neural networks learn which part of point clouds are walkable.
Although the image representation can also have a fourth channel as distance information, the scaling operation can make some important obstacles or free points disappear, which changes the structure of the map.
Secondly, the number of pixels in an image increases exponentially as the size of the image increase.
The number of points in point clouds equals the number of pixels that represent an obstacle or frontier in a map, which is not an exponential relation unless all the pixels in a map are obstacles or frontiers.

\section{Conclusions And Future Work}
\label{sec:conclude}

In this paper, we present a novel state representation method using 4D point-clouds-like information and design the framework to learn an efficient exploration strategy.
Our proposed method can solve the problems that come with using images as observations.
The experiments demonstrate the effectiveness of our proposed method, compared with other five commonly used methods.
For the future work, other network structures and RL algorithms can be modified and applied to the robot exploration problem with point clouds as input.
The converge speed of training may also be improved by optimizing the training techniques.
Besides, the multi-robot exploration problem may also use point clouds to represent the global information.

\addtolength{\textheight}{-8cm}   



\bibliographystyle{IEEEtran}
\bibliography{ifacconf}

\end{document}